\documentclass[12pt]{article}
\usepackage{amsmath}
\usepackage{graphicx}
\usepackage{enumerate}
\usepackage{natbib}
\usepackage{url} 

\usepackage{bibentry}

\usepackage{diagbox}

\usepackage{tikz}
\usetikzlibrary{shapes, arrows}
\usepackage{multirow}
\usepackage{amssymb}
\usepackage{mathtools}
\usepackage{amsthm}
\usepackage{microtype}
\usepackage{graphicx}

\usepackage{subcaption}

\usepackage{booktabs}

\usepackage{float}
\theoremstyle{plain}
\newtheorem{theorem}{Theorem}

\newtheorem{lemma}{Lemma}

\theoremstyle{definition}

\theoremstyle{remark}

\usepackage{bibentry}
\usepackage{color}

\usepackage{mathrsfs}
\usepackage{comment}

\begin{document}

\def\spacingset#1{\renewcommand{\baselinestretch}%
{#1}\small\normalsize} \spacingset{1}

\title{\bf Learning Optimal Distributionally Robust Individualized Treatment Rules Integrating Multi-Source  Data}

\author{Wenhai Cui$^1$, Wen Su$^2$ and Xingqiu Zhao$^1$\\
$^1$Department of Applied Mathematics, The Hong Kong Polytechnic University\\ 
$^2$Department of Biostatistics, City University of Hong Kong\\ 
}

\maketitle

\bigskip

\begin{abstract}
Integrative analysis of multiple datasets for estimating optimal individualized treatment rules (ITRs) can enhance decision efficiency.
A central challenge is posterior  shift, wherein the conditional distribution of potential outcomes given covariates differs between source and target populations.
 We propose a prior information-based distributionally robust ITR (PDRO-ITR) that maximizes the worst-case policy value over a covariate-dependent distributional uncertainty set, {ensuring robust performance under posterior  shift.}
The uncertainty  set is constructed as an individualized combination of source distributions, with weights combining prior source-membership probabilities
and deviation terms constrained to the probability simplex to accommodate posterior shift. 
We derive a closed-form solution for the PDRO-ITR and develop an adaptive procedure to tune the uncertainty level. 
We establish risk bounds for the PDRO-ITR estimator, which guarantees  robust performance under the worst case. Extensive simulations and two real-data applications demonstrate that the proposed method achieves superior performance  compared to existing approaches.
\end{abstract}

\vfill

\newpage
\spacingset{1.9} 

\section{Introduction}

Individualized treatment rules  have gained  widespread applications in  public policy design and precision medicine. A classic method for estimating the  optimal ITR  is  quality learning (Q-learning).  Q-learning estimates the conditional expectation of potential outcome given treatment and covariates \citep{qian2011performance, song2015penalized, shi2018high}, and then selects  the treatment that maximizes  the conditional expectation.  An alternative is policy search, which directly maximizes an estimated policy value over a pre-specified policy class \citep{zhao2012estimating, zhang2012robust}. 
In medical applications, electronic health records often pool data from multiple source domains that exhibit outcome heterogeneity, raising concerns about the generalizability of learned ITRs to target populations.

Distributional shifts can arise for several reasons. First, the training data may not adequately represent the target population due to design limitations. For example, women are often underrepresented in HIV research \citep{curno2016systematic}.
In AIDS Clinical Trials Group Study 175, female participants  accounted for less than $18 \%$ of the  sample, and in the Oregon Health Insurance Experiment, racial minority participants accounted for less than $23.69 \%$ of the study sample. Second, the treatment environment may differ across settings due to variations in healthcare infrastructure, quality of care, and unobserved confounding factors, leading to changes in potential outcomes. 
Third, the target distribution itself may evolve over time. For instance, in policy design, income is often the primary outcome, and its distribution may shift with labor market dynamics. Consequently, it remains a critical challenge to develop ITRs that perform robustly across underrepresented populations, heterogeneous environments, and temporally evolving settings.

There exists limited research on distributional shift in the context of ITRs.  {
Specifically, 
addressing on individual heterogeneity, \cite{Cui20102025} proposed a minimax policy targeting the conditional quantile of individual treatment effects, where the uncertainty set is restricted to a class of distributions parametrized by pre-specified Bernstein copulas.}
Focusing on covariate shift, \cite{mo2021learning} developed an ITR that maximizes the worst-case policy value within an uncertainty set of distributions that are close to the source empirical distribution in Kullback–Leibler divergence. However, these approaches consider single-source settings.
In the context of multi-source heterogeneity, \cite{shi2018maximin}   proposed to estimate a linear ITR by optimizing the worst-case performance over the discrete set of available source distributions.
Recently, \cite{zhang2024minimax} proposed a minimax estimator for the conditional average treatment effect (CATE)  by minimizing the worst-case regret, defined as the mean squared error between the true and estimated CATE, across underlying distributions expressible as linear combinations of multiple source distributions.
However, minimizing worst-case CATE regret is not equivalent to maximizing policy value, and thus does not directly target optimal decision-making. Furthermore, the uncertainty sets employed in these methods do not incorporate  prior information, such as dependence between covariates and group indicators, and may be overly large, leading to excessively conservative ITRs. 
In the linear regression setting, \cite{zhan2024domain} constructed a less conservative uncertainty set by leveraging covariate similarity between source and target domains, and proposed minimizing the worst-case prediction error over this set. Their approach, however, requires a sufficiently large number of unlabeled samples from the target population and focuses on the prediction task  for linear models  rather than decision-making. {Moreover, obtaining a  distributionally robust decision rule under an uncertainty set poses additional computational challenges: the classical policy value function \citep{zhao2012estimating} is non-smooth and non-concave, making the worst-case value maximization problem fundamentally nontrivial.}

In this paper, we develop a distributionally robust ITR  under posterior shift,   {where the optimal ITR for the target population may vary across sites, time or other unobserved confounders. In such settings, only a limited amount of unlabeled target data is initially available, and treatments for new target individuals must be recommended with caution. However, the limited information about the target distribution makes it challenging to obtain a reliable ITR estimator.} To ensure robustness under distributional uncertainty,
we begin with constructing a prior information–based uncertainty set, defined as individualized weighted combination of source distributions. The individualized weights combine prior source-membership probabilities, i.e., the conditional probabilities that an individual originates from each source, with distributionally robust deviation terms that capture the direction of posterior shift. 
{We then maximize the worst-case policy value over the  uncertainty set, yielding the proposed prior-information–based distributionally robust ITR (PDRO-ITR). }
When a small calibration sample from the target distribution is available, we further introduce an adaptive procedure that calibrates the uncertainty level of the set.

Our contributions are summarized as follows:
\begin{itemize}
   \item[(i)] {\it Strong {robustness }:} 
   The uncertainty set contains a broad class of distributions, including all available source distributions  as well as linear combinations of the source distributions.  By maximizing the worst-case policy value over the uncertainty set, our framework ensures robust performance across  a wide range of plausible distributions.

\item[(ii)] {\it Flexible decision robustness–efficiency trade-off:} 
        {In the uncertainty set, the individualized weights  represent the prior probability of an individual originating from a given source. By varying the uncertainty level, the distributions in the set interpolate between individualized weighted combinations of the source distributions and simple linear combinations of the source distributions. An adaptive tuning procedure controls the uncertainty level,  preventing overly ``conservative'' ITR estimation while maintaining robustness.}
 
   \item[(iii)] {\it Computationally efficient implementation:} 
 We derive a closed-form of PDRO-ITR whose decision function is an individualized, covariates-dependent weighting of source CATEs. Rather than solving a max-min optimization problem, our method only requires  estimating the weighting function and source CATEs using existing machine-learning tools.
                
   \item[(iv)] {\it Theoretical guarantees and empirical superior performance:} 
   We establish risk bounds for PDRO-ITR, providing rigorous performance guarantees under distributional shift. Extensive simulations and real-data applications demonstrate the superiority of our method relative to existing methods.
\end{itemize}
The remainder of this paper is organized as follows. Section 2 introduces key definitions, including the value function and the optimal ITR under the target distribution. Section 3 presents the proposed methodology for estimating the PDRO-ITR. Section 4 provides the estimation procedure, and Section 5 establishes theoretical results. Section 6 reports simulation results, while Section 7 illustrates the proposed method by analyzing the AIDS Clinical Trials Group Study 175 dataset and the Oregon Health Insurance Experiment dataset. Finally, Section 8 concludes with a discussion and directions for future research.

\section{Preliminaries}
Consider  source domains $\{\boldsymbol{X}^{(s)}, A^{(s)}, Y^{(s)}\}_{s=1}^{|\mathcal{S}|}$ collected from different sub-populations, indexed by $s \in \mathcal{S} = \{1, 2, \ldots, |\mathcal{S}|\}$, where $|\mathcal{S}|$ denotes the cardinality of set $\mathcal{S}$. The distribution  of source domains may vary by experimental geographic location or demographic characteristics such as race. The covariate vector $\boldsymbol{X}^{(s)}\in \mathcal{X}$ is  a $p$-dimensional  vector.  The binary treatment is denoted by $A^{(s)} \in \mathcal{A} = \{1, 0\}$. Let $Y^{(s)}(1)$ and $Y^{(s)}(0)$ denote  potential  outcomes. 
For the $s$th source  domain, we denote the distributions  of covariates and   potential outcomes  as
$$\boldsymbol{X}^{(s)} \sim \mathbb{P}_{X}^{(s)},  \quad \quad\quad \{Y^{(s)}(1),Y^{(s)}(0)\}  \mid \boldsymbol{X}^{(s)}\sim  \mathbb{P}_{Y(1),Y(0) \mid \boldsymbol{X} }^{(s)},$$
where $\mathbb{P}_{\boldsymbol{X}}^{(s)}$ denotes the marginal distribution function of ${\boldsymbol{X}^{(s)}}$ and $ \mathbb{P}_{Y(1),Y(0) \mid \boldsymbol{X} }^{(s)}$ denotes the conditional distribution function of  potential outcomes $\{Y^{(s)}(1),Y^{(s)}(0)\} $ given ${\boldsymbol{X}^{(s)}}$. 
We assume that for each $s \in \mathcal{S}$, the observed outcome is given by $Y^{(s)}=Y^{(s)}(1)A^{(s)}+Y^{(s)}(0)(1-A^{(s)})$ and the potential outcomes ${Y^{(s)}(0), Y^{(s)}(1)}$ are independent of the treatment  $A^{(s)}$ given the covariates $\boldsymbol{X}^{(s)}$ \citep{rosenbaum1983central, robins2000marginal}.

For the target domain, we write  the distribution functions   of  covariates and potential outcomes  as 
$$\boldsymbol{X}^{(t)} \sim \mathbb{P}_{\boldsymbol{X}}^{(t)},  \quad \quad\quad \{Y^{(t)}(1),Y^{(t)}(0)\}  \mid \boldsymbol{X}^{(t)} \sim  \mathbb{P}_{Y(1),Y(0) \mid \boldsymbol{X} }^{(t)}.$$
We focus on posterior shift, where $ \mathbb{P}_{Y(1),Y(0) \mid \boldsymbol{X} }^{(t)}$ and  $\mathbb{P}_{Y(1),Y(0) \mid \boldsymbol{X} }^{(s)}$  may differ.  
Consider the setting where labeled outcomes in the target domain are unavailable, which poses challenges for generalizing ITRs from source domains to the target population. This situation commonly arises when outcome labeling is costly or when an ITR must be deployed at a new site or for a new  population.

The  ITR $d$ is a function  mapping from space {$\mathcal{X}$} to space $\mathcal{A}$. Correspondingly, the potential outcome under a policy $d$ is given by
\[
Y^{(t)}(d) = Y^{(t)}(1) \mathbb{I}\left\{d(\boldsymbol{X}^{(t)}) = 1\right\} + Y^{(t)}(0)  \mathbb{I}\left\{d(\boldsymbol{X}^{(t)}) = 0\right\},
\]  
where $\mathbb{I}$ is an indicator function.
Furthermore, the policy value  for ITR  $d$ is defined as
$$
\mathcal{V}(d; \mathbb{P}^{(t)})=E_{\mathbb{P}^{(t)}}\left\{Y^{(t)}(d)\right\},
$$
where $\mathbb{P}^{(t)}$ is the joint distribution of $\{\boldsymbol{X}^{(t)},Y^{(t)}(1),Y^{(t)}(0)\}$.
The conditional average treatment effect (CATE) is defined as $$C\left\{\boldsymbol{x}; \mathbb{P}_{Y(1), Y(0)|\boldsymbol{X}}^{(t)}\right\}=E_{ \mathbb{P}_{Y(1), Y(0)|\boldsymbol{X}}^{(t)}}\left\{Y^{(t)}(1)-Y^{(t)}(0) \mid \boldsymbol{X}^{(t)}=\boldsymbol{x}\right\},$$ where the expectation is taken over the conditional distribution $\mathbb{P}_{Y(1), Y(0)|\boldsymbol{X}}^{(t)}$. 
Then,
the policy value can be reformulated as
\begin{equation}
\begin{aligned}
\mathcal{V}(d, \mathbb{P}^{(t)})-E_{\mathbb{P}^{(t)}}[Y(0)]
=&E_{\mathbb{P}^{(t)}}[Y\big(d(\boldsymbol{X})\big)-Y(0)]\\
=&E_{\mathbb{P}^{(t)}}[\left\{Y(1)-Y(0)\right\}d(\boldsymbol{X})]\\
=&E_{\mathbb{P}_{\boldsymbol{X}}^{(t)}}\left[C\left\{\boldsymbol{X}; \mathbb{P}_{Y(1), Y(0)|\boldsymbol{X}}^{(t)}\right\}d(\boldsymbol{X})\right]
,
\nonumber
\end{aligned}
\end{equation}
where the second equality follows from $$Y^{(t)}\big(d(\boldsymbol{X}^{(t)})\big)=Y^{(t)}\big(0\big)+\left\{Y^{(t)}(1)-Y^{(t)}(0)\right\}d(\boldsymbol{X}^{(t)}).$$  To simplify notation, for any functions $g_1$ and $g_2$,  denote $E_{\mathbb{P}^{(t)}}\left\{g_1\big(\boldsymbol{X},Y(1),Y(0)\big)\right\}=E_{\mathbb{P}^{(t)}}\left\{g_1\big(\boldsymbol{X}^{(t)},Y^{(t)}(1),Y^{(t)}(0)\big)\right\}$ and $E_{\mathbb{P}_{\boldsymbol{X}}^{(t)}}\left\{g_2\big(\boldsymbol{X}\big)\right\}=E_{\mathbb{P}_{\boldsymbol{X}}^{(t)}}\left\{g_2\big(\boldsymbol{X}^{(t)}\big)\right\}$.  
Consequently, maximizing the policy value is equivalent to the following optimization problem:
\begin{equation}
\begin{aligned}
\underset{{d}}{\max } \ E_{\mathbb{P}_{\boldsymbol{X}}^{(t)}}\left[C\left(\boldsymbol{X}; \mathbb{P}_{Y(1), Y(0)|\boldsymbol{X}}^{(t)}\right)d(\boldsymbol{X})\right].
\nonumber
\end{aligned}
\end{equation}
This problem admits a closed-form solution $ \mathbb{I}\left\{C\left(\boldsymbol{X}; \mathbb{P}_{Y(1), Y(0)|\boldsymbol{X}}^{(t)}\right)>0\right\}$. 

\section{Methodology}

\subsection{Distributionally robust ITRs}
Since the conditional distribution of potential outcomes 
$\mathbb{P}_{Y(1),Y(0) \mid \boldsymbol{X} }^{(t)}$ may differ from  source distributions $\mathbb{P}_{Y(1),Y(0) \mid \boldsymbol{X} }^{(s)}$  for $s\in \mathcal{S}$,  the ITR estimated based on source data cannot perform well in the target domain.
To address this challenge, we consider an uncertainty  set of conditional  distributions that combines information from the observed source domains: $$\mathcal{U}_0=\left\{ \mathbb{T}_{Y(1),Y(0) \mid \boldsymbol{X}} \quad \Bigg| \quad \mathbb{T}_{Y(1),Y(0) \mid \boldsymbol{X}}= \sum_{s=1}^{|\mathcal{S}|}\rho_s \mathbb{P}_{Y(1),Y(0) \mid \boldsymbol{X} }^{(s)},  \sum_{s=1}^{|\mathcal{S}|}\rho_s=1, \rho_s \geq0 \right\}.$$ 
The set $\mathcal{U}_0$ contains all possible linear mixtures of the source distributions. Then,
the distributionally robust ITR (DRO-ITR) is  defined as 
\begin{equation}
\label{e_dro}
\begin{aligned}
d^{*}_{dro}= \underset{{d}}{\arg \max }\min_{\mathbb{T}_{Y(1), Y(0)|\boldsymbol{X}} \in \mathcal{U}_0} E_{\mathbb{P}_{\boldsymbol{X}}^{(t)}}\left[C\left\{\boldsymbol{X};  \mathbb{T}_{Y(1), Y(0)|\boldsymbol{X}}\right\}d(\boldsymbol{X})\right].
\end{aligned}
\end{equation}
The formulation ensures that DRO-ITR performs well even under the  worst-case distribution over the uncertainty class $\mathcal{U}_0$. However, directly solving the max–min problem in Equation \eqref{e_dro}  is computationally  challenging. The following theorem provides an explicit form of the distributionally robust ITR.
\begin{theorem} Let $\mathcal{P}=\{ \boldsymbol{\rho}=(\rho_1, \ldots,\rho_{|\mathcal{S}|}) \quad|\quad \sum_{s=1}^{|\mathcal{S}|}\rho_s=1, \rho_s \geq0\} $. The distributionally robust ITR is
    $d^{*}_{dro}= \mathbb{I}(\sum_{s=1}^{|\mathcal{S}|} \rho_{s}^{*} C\{\boldsymbol{X}; \mathbb{P}_{Y(1), Y(0)|\boldsymbol{X}}^{(s)}\}>0),$  where  $\boldsymbol{\rho}^{*}=(\rho_{1}^{*}, \rho_{2}^{*}, \ldots, \rho_{|\mathcal{S}|}^{*})$   is given by
 $$\boldsymbol{\rho}^{*}=
    \arg \min_{\boldsymbol{\rho}\in \mathcal{P}}
    E_{\mathbb{P}_{\boldsymbol{X}}^{(t)}} \left[
    \sum_{s=1}^{|\mathcal{S}|} \rho_sC\left(\boldsymbol{X};  \mathbb{P}_{Y(1), Y(0)|\boldsymbol{X}}^{(s)}\right)\mathbb{I}\left\{\sum_{s=1}^{|\mathcal{S}|} \rho_sC\left(\boldsymbol{X};  \mathbb{P}_{Y(1), Y(0)|\boldsymbol{X}}^{(s)}\right)>0\right\}\right].$$
\end{theorem}
By Theorem 1, the DRO-ITR  corresponds to a linear combination of source CATEs.
\subsection{Distributionally robust ITR with prior information}
\subsubsection{Motivating Example}
To account for potential heterogeneity of target population, we introduce a subpopulation variable $S^{(t)} \in \mathcal{S}$ that characterizes subpopulation structure.  
Specifically, we assume that the conditional distribution of the potential outcomes given covariates and subpopulation index satisfies
$$\{Y^{(t)}(1),Y^{(t)}(0)\}  \mid (\boldsymbol{X}^{(t)}={\boldsymbol{x},} {S}^{(t)}=s) \sim  \mathbb{P}_{Y(1),Y(0) \mid \boldsymbol{X}=\boldsymbol{x} }^{(s)}.$$ 
Consequently, if  $S^{(t)}\equiv s$, the target conditional distribution  $\mathbb{P}_{Y(1),Y(0) \mid \boldsymbol{X}=\boldsymbol{x}}^{(t)}$ is identical to  the $s$th source  conditional distribution $\mathbb{P}_{Y(1),Y(0)| \boldsymbol{X}=\boldsymbol{x}}^{(s)}$.
Consider the complete data structure in the target domain, $\{\boldsymbol{X}^{(t)}, A^{(t)}, Y^{(t)}(0), Y^{(t)}(1), S^{(t)}\}$, where $S^{(t)}$ is unobserved.
Then, marginalizing over $S^{(t)}$ yields  \begin{equation}
\begin{aligned}
\label{eq:latent-mixture}
\mathbb{P}_{Y(1),Y(0) \mid \boldsymbol{X}=\boldsymbol{x}}^{(t)}=\sum_{s\in \mathcal{S}}\mathbb{P}(S^{(t)}=s|\boldsymbol{X}^{(t)}=\boldsymbol{x})\mathbb{P}_{Y(1),Y(0)| \boldsymbol{X}=\boldsymbol{x}}^{(s)}.
\end{aligned}
\end{equation}
Crucially, the mixing weights $\mathbb{P}(S^{(t)}=s|\boldsymbol{X}^{(t)})$ in \eqref{eq:latent-mixture} are {covariate-dependent}. Consequently, the true target distribution likely lies outside the baseline uncertainty set $\mathcal{U}_0$, which assumes constant weights $\boldsymbol{\rho}$. This limitation motivates the construction of a more flexible uncertainty set.

\subsubsection{Prior information-based distributionally robust ITR}
Directly estimating  $\mathbb{P}(S^{(t)}=s|\boldsymbol{X}^{(t)}=\boldsymbol{x})$ is infeasible due to the absence of observed $ S^{(t)}$ in the target domain.  Moreover, the mixture distribution assumption in Equation (\ref{eq:latent-mixture}) may not hold exactly in practice.
However, this motivates us to leverage the prior information from the source domain to construct an uncertainty set that covers  the conditional distribution $\mathbb{P}_{Y(1),Y(0) \mid \boldsymbol{X}}^{(t)}$.

We rewrite the complete source population  as 
$\{\boldsymbol{X}, A, Y(0),Y(1), S\}$, where $S=s$ indicates that the observation comes from the $s$th source domain.
Let $\omega^{(s)}_0(\boldsymbol{x})=\mathbb{P}(S=s|\boldsymbol{X}=\boldsymbol{x})$ for $s \in \mathcal{S}$.
Using these covariate-dependent weights, we define an extended uncertainty set of conditional outcome distributions: {\small
$$\mathcal{U}_1(\delta)=\left\{ \mathbb{T}_{Y(1),Y(0) \mid \boldsymbol{X}}   \mid \mathbb{T}_{Y(1),Y(0) \mid \boldsymbol{X}}= \sum_{s=1}^{|\mathcal{S}|}  \left\{ \delta\omega^{(s)}_0(\boldsymbol{X})+(1-\delta)\rho_s \right\}\mathbb{P}_{Y(1),Y(0) \mid \boldsymbol{X} }^{(s)}, \boldsymbol{\rho}\in \mathcal{P}
\right\},$$ }
where   $\delta \in [0,1]$ and $ \mathcal{P}=\{(\rho_1, \ldots, \rho_{|\mathcal{S}|})  |\sum_{s=1}^{|\mathcal{S}|}\rho_s=1, \rho_s \geq0 \}$.
This formulation interpolates between two extremes: (i) a fully prior information-based specification weighted by $\{\omega^{(s)}_0(\boldsymbol{x})\}_{s \in\mathcal{S}}$ ;
(ii) an  uncertain mixture over all possible class proportions $\boldsymbol{\rho} \in \mathcal{P}$. The covariate-dependent
 weights $\{\omega^{(s)}_0(\boldsymbol{x})\}_{s \in\mathcal{S}}$  determine how strongly each source distribution $\mathbb{P}_{Y(1),Y(0)| \boldsymbol{X}}^{(s)}$  contributes to the target  distribution $\mathbb{P}_{Y(1),Y(0)| \boldsymbol{X}}^{(t)}$, 
while the parameters $\{\rho_s\}_{s \in\mathcal{S}}$ quantify potential deviations from the source distributions.  
The mixing parameter \(\delta\) governs the degree of reliance on source-domain information: higher \(\delta\) places more trust in the source distributions, whereas lower \(\delta\) allows greater flexibility to account for distributional uncertainty. Therefore,
the uncertainty class
$\mathcal{U}_1(\delta)$ encompasses a flexible family of conditional distributions.

The prior information-based distributionally robust ITR (PDRO-ITR) under the uncertainty  class is defined as
\begin{equation}
\label{e4}
\begin{aligned}
d^{*}_{pdro}(\boldsymbol{X}; \delta)= \underset{{d}}{\arg \max }\min_{\mathbb{T}_{Y(1), Y(0)|\boldsymbol{X}} \in \mathcal{U}_1(\delta)} E_{\mathbb{P}_{\boldsymbol{X}}^{(t)}}\left[C\left\{\boldsymbol{X};  \mathbb{T}_{Y(1), Y(0)|\boldsymbol{X}}\right\}d(\boldsymbol{X})\right].
\end{aligned}
\end{equation}
Crucially, the class $\mathcal{U}_1(\delta)$   anchors the uncertainty to the prior information-based  weights $\{\omega^{(s)}_0(\boldsymbol{x})\}_{s \in\mathcal{S}}$, avoiding overly conservative  decision rules induced by max-min framework.

The following theorem provides a closed-form characterization of the PDRO-ITR, avoiding the need to directly solve the original max–min optimization problem.
\begin{theorem} Define  $C^{(s)}_0(\boldsymbol{x})=C\left\{\boldsymbol{x};  \mathbb{P}_{Y(1), Y(0)|\boldsymbol{X}=\boldsymbol{x}}^{(s)}\right\}$ for $s \in \mathcal{S}$, and  let the nuisance parameter be 
 $\eta_0 =\{\omega^{(1)}_0,\ldots \omega^{(|\mathcal{S}|)}_0, C_0^{(1)},\ldots, C^{(|\mathcal{S}|)}_0\}$.
For \(\boldsymbol{\rho} \in \mathcal{P}\) and \(\delta\in [0,1] \), define
 $
f_{\boldsymbol{\rho},\eta_0}^{\delta}(\boldsymbol{x})=\sum_{s=1}^{|\mathcal{S}|}\left\{ \delta {\omega}^{(s)}_0(\boldsymbol{x})+(1-\delta)\rho_s \right\}C^{(s)}_0\left(\boldsymbol{x}\right)
$.  Then, the PDRO-ITR is
    $$ d^{*}_{pdro}(\boldsymbol{X}; \delta)=\mathbb{I}\left[f_{\boldsymbol{\rho}_0^{\delta},\eta_0}^{\delta}(\boldsymbol{X})>0\right],$$  where  $\boldsymbol{\rho}_0^{\delta}=(\rho_{0,1}^{\delta}, \ldots, \rho_{0,|\mathcal{S}|}^{\delta})$  is given by
 $$\boldsymbol{\rho}_0^{\delta}=
    \arg \min_{\boldsymbol{\rho}\in \mathcal{P}}
    E_{\mathbb{P}_{\boldsymbol{X}}^{(t)}}
\left[f_{\boldsymbol{\rho},\eta_0}^{\delta}(\boldsymbol{X})
\mathbb{I}\left\{f_{\boldsymbol{\rho},\eta_0}^{\delta}(\boldsymbol{X})>0\right\}\right].$$
\end{theorem}
Define the individualized weight functions $\mathcal{W}_{s}(\boldsymbol{X}, \rho_{0, s}^{\delta}, {\delta})= \delta {\omega}^{(s)}_0(\boldsymbol{X})+(1-\delta)\rho_{0, s}^{\delta}$ for $s \in \mathcal{S}$. By Theorem 1, the PDRO-ITR assigns treatment according to an individualized weighted sum of the source CATEs $\{C^{(s)}_0\}_{s\in\mathcal{S}}$, where the individualized weight functions $\{\mathcal{W}_{s}\}_{s \in \mathcal{S}}$ capture both covariate-based source relevance and the robustness adjustment.

Given  $\boldsymbol{X}=\boldsymbol{x}$ and $\delta$, the PDRO-ITR considers the worst case over the prior information-based surface, i.e.,  $$\widetilde{\mathcal{P}}(\boldsymbol{x},\delta)=\left[\left(\mathcal{W}_{1}(\boldsymbol{x}, \rho_1, {\delta}), \mathcal{W}_{2}(\boldsymbol{x}, \rho_s, {\delta}), \ldots ,\mathcal{W}_{|\mathcal{S}|}(\boldsymbol{x}, \rho_{|\mathcal{S}|}, {\delta}) \right) \mid  \sum_{s=1}^{|\mathcal{S}|}\rho_s=1, \rho_s \geq0\right]. $$ Clearly,  it follows that $\widetilde{\mathcal{P}}(\boldsymbol{x},\delta)\subseteq \mathcal{P}$  for any $\boldsymbol{x}\in \mathcal{X}$ and $\delta \in [0,1]$.  
The parameter $\delta$ controls the size of the surface  $\widetilde{\mathcal{P}}$ while the covariates $\boldsymbol{x}$ determines its center. 

As illustrated in Figure 1, the prior information-based surface $\widetilde{\mathcal{P}}(\boldsymbol{x},\delta)$ covers  the true curve  defined by the conditional probabilities, i.e., $$\left(\mathbb{P}(S^{(t)}=1|{X}^{(t)}={x}), \mathbb{P}(S^{(t)}=2|{X}^{(t)}={x}), \mathbb{P}(S^{(t)}=3|{X}^{(t)}={x})\right),$$ but occupies a smaller region than the  simplex surface $\mathcal{P}$.  Intuitively, restricting the surface in this way prevents the ITR from being overly conservative.
\begin{figure}[h]	
	\centering
	\includegraphics[width=0.6\linewidth]{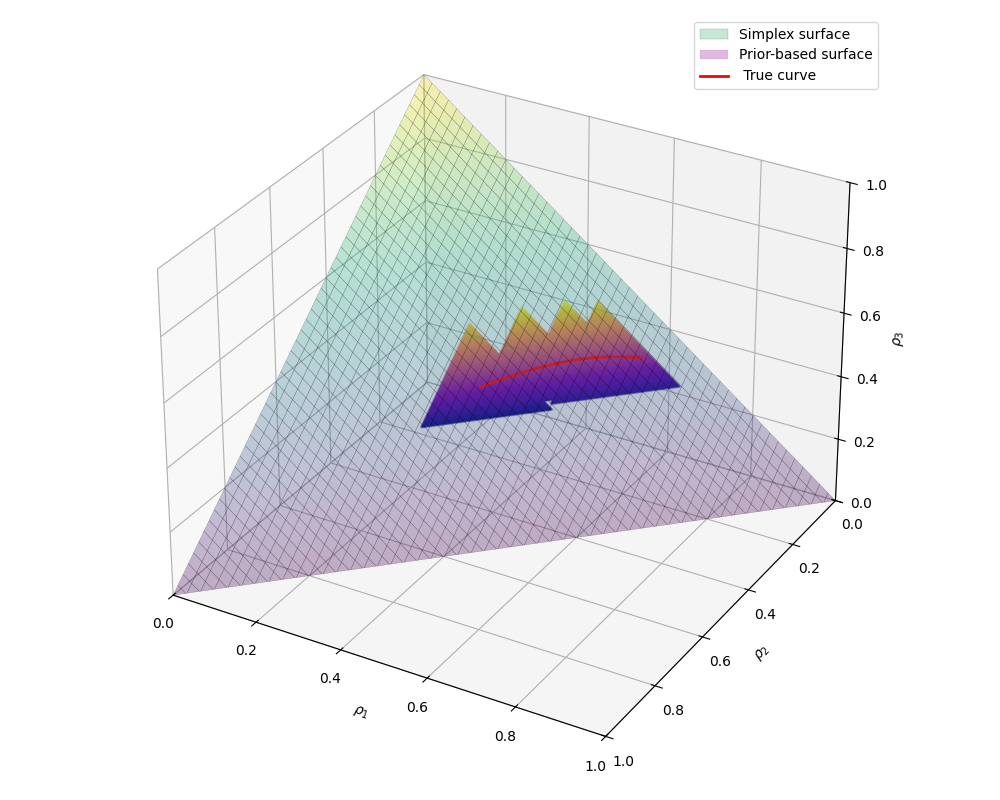}
        \caption{Geometric illustration of the simplex surface \(\mathcal{P}\) and the prior information-based surfaces  
       $\widetilde{\mathcal{P}}(\boldsymbol{x},\delta)$ evaluated at \(x = 0, ({1}/{3}), ({2}/{3}), 1\) with \(\delta = 0.8\); The weight functions in the $\widetilde{\mathcal{P}}(\boldsymbol{x},\delta)$ are defined as
$
\omega^{(s)}_{0}( {x})= {\exp(\beta_s x)}/\{\sum_{s \in \mathcal{S}} \exp(\beta_s x)\}
$ for $s=1, 2, 3$,
where \((\beta_1, \beta_2, \beta_3) = (1, -1, 0.5)\); The true curve is defined as $\left[\mathbb{P}(S^{(t)}=1|{X}^{(t)}={x}), \mathbb{P}(S^{(t)}=2|{X}^{(t)}={x}), \mathbb{P}(S^{(t)}=3|{X}^{(t)}={x})\right],$ 
and we assume that
\( \mathbb{P}(S^{(t)}=s|{X}^{(t)}={x}) = \omega^{(s)}_{0}( {x})\), i.e., the source and target domains share the same subgroup structure.
        }
\end{figure}

\section{Estimation procedure for PDRO-ITR}
Suppose that we observe  samples $\{(X_i^{(s)}, A_i^{(s)}, Y_i^{(s)})\}_{i=1}^{N_s}$ for each source domain and  unlabeled covariates $\{X_i^{(t)}\}_{i=1}^{N_t}$ for the target domain.  Let $\mathcal{I}_1^{(s)} = \{i: A_i^{(s)} = 1\}$ and $\mathcal{I}_0^{(s)} = \{i: A_i^{(s)} = 0\}$ denote the index sets for the treated and control groups in the $s$-th  source, respectively.
For theoretical convenience, we assume equal sample sizes for all domains, i.e., $N_t=N_s=n/(|\mathcal{S}|+1)$  for each $s\in \mathcal{S}$.

\noindent \textbf{Step 1: Estimation of CATEs Using Deep Neural Networks.}

A variety of methods can be used to estimate CATEs \citep{wager2018estimation, yang2020improved}. 
For each source domain $s \in \mathcal{S}$, we separately estimate  the conditional expectations $E\left\{Y^{(s)}\mid \boldsymbol{X}^{(s)}=\boldsymbol{x}, A^{(s)}=1\right\}$ and $E\left\{Y^{(s)}\mid \boldsymbol{X}^{(s)}=\boldsymbol{x}, A^{(s)}=0\right\}$ using feedforward neural networks. Specifically, the estimators are obtained by minimizing the empirical mean squared error, i.e., {\small
\[
\widehat{f}_{1}^{(s)}= \arg\min_{ f\in\mathcal{F}_0} \frac{1}{|\mathcal{I}_1^{(s)}|}\sum_{i\in \mathcal{I}_1^{(s)}}\{f(X_i^{(s)}) - Y_i^{(s)}\}^{2}, 
\widehat{f}_{0}^{(s)}=\arg\min_{f\in\mathcal{F}_0}\frac{1}{|\mathcal{I}_0^{(s)}|}\sum_{i\in \mathcal{I}_0^{(s)}}\{f(X_i^{(s)}) - Y_i^{(s)}\}^{2}, 
\]}
 where $ \mathcal{F}_0$ denotes the class of feedforward neural networks (FNNs) with the rectified linear unit (ReLU) activation function, $\max(0,u)$. The estimated source-specific CATEs are given by
$
\widehat{C}^{(s)}(\boldsymbol{X}) = \widehat{f}_{1}^{(s)}(\boldsymbol{X}) - \widehat{f}_{0}^{(s)}(\boldsymbol{X}),
$ for $s \in \mathcal{S}$.

\noindent \textbf{Step 2: Estimation of $\omega^{(s)}_0(\boldsymbol{x}) = \mathbb{P}(S = s \mid \boldsymbol{X} = \boldsymbol{x})$}.

We estimate  $\mathbb{P}(S = s \mid \boldsymbol{X} = \boldsymbol{x})$ using the multinomial logistic regression over a function class $\mathcal{F}_1=\left\{{\omega^{(s)}(\boldsymbol{x})=\exp \boldsymbol{\beta}_s}^{T}\boldsymbol{x}/\sum_{j\in \mathcal{S}}{\exp \boldsymbol{\beta}_j}^{T}\boldsymbol{x}: \boldsymbol{\beta}_j\in \mathbb{R}^{p} 
 \text{ for } j \in \mathcal{S}\right\}$. The details are shown in Section 8.1.1 of  \cite{david2000applied}.
Let $\widehat{\omega}^{(s)}(\boldsymbol{x})$ for $s\in \mathcal{S}$ denote the estimated conditional probability.
Then, we obtain the individualized weight functions
$\widehat{\mathcal{W}}_s(\boldsymbol{x},\rho_s,\delta)= \delta\widehat{\omega}^{(s)}(\boldsymbol{x})+(1-\delta)\rho_s.$

\noindent \textbf{Step 3: Estimation of PDRO-ITR}

\begin{figure}[h]	
	\centering
	\includegraphics[width=0.6\linewidth]{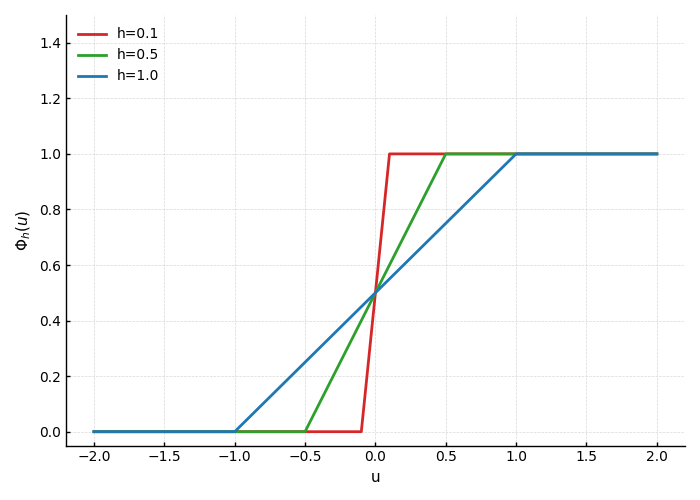}
        \caption{ Plot of the smoothed surrogate function $\Phi_h(u)$ under varying $h$.}
\end{figure}

{
In the setting of posterior shift with invariant marginal covariate distributions, we utilize the pooled source  and target covariates, denoted by $\{\boldsymbol{X}_i\}_{i=1}^{n}$, as a representative sample of the target  population.} 
Then, we
 estimate $\boldsymbol{\rho}_0^{\delta}$ by minimizing a smoothed surrogate loss 
$$
\widehat{\boldsymbol{\rho}}^{\delta}_{n}= \arg \min_{\boldsymbol{\rho}\in \mathcal{P}}
 \frac{1}{n} \sum_{i=1}^n
\left[\sum_{s=1}^{|\mathcal{S}|}\widehat{\mathcal{W}}_s\left(\boldsymbol{X}_{i}, \rho_s,\delta\right)\widehat{C}^{(s)}\left(\boldsymbol{X}_{i}\right)
\Phi_{h}\left\{\sum_{s=1}^{|\mathcal{S}|}\widehat{\mathcal{W}}_s\left(\boldsymbol{X}_{i}, \rho_s,\delta\right)\widehat{C}^{(s)}\left(\boldsymbol{X}_{i}\right)\right\}\right],$$
where $\widehat{\boldsymbol{\rho}}^{\delta}_{n}=(\widehat{\rho}^{\delta}_{n,1}, \widehat{\rho}^{\delta}_{n,2}, \ldots, \widehat{\rho}^{\delta}_{n,|\mathcal{S}|})$  and  $\Phi_h(u)=I(|u|\leq h)\frac{u+h}{2h}+I(u>h)$  
and $h > 0$. The function $\Phi_h$ serves as a smooth, differentiable surrogate for indicator function $\mathbb{I}$. As shown in Figure 2,  the hyper-parameter $h$ controls the smoothness of  $\Phi_h$.

To computationally solve this problem while ensuring that $\boldsymbol{\rho}$ lies in the probability simplex $\mathcal{P} = \{ \boldsymbol{\rho} \in \mathbb{R}^{|\mathcal{S}|}: \sum_{s} \rho_s = 1, \rho_s \geq 0 \}$, we employ the softmax parameterization. Specifically, let $\rho_s = \exp(z_s) / \sum_{j=1}^{|\mathcal{S}|} \exp(z_j)$ for $s \in \mathcal{S}$, where $\boldsymbol{z} \in \mathbb{R}^{|\mathcal{S}|}$ denotes the unconstrained parameters. 
This formulation allows us to perform unconstrained optimization over $\boldsymbol{z}$ using the adaptive moment estimation (Adam) algorithm \citep{adam}.

\noindent \textbf{Step 4: Tuning hyperparameter  $\delta$.}

When some information about the target distribution is available, we propose an effective method for tuning the hyperparameter $\delta$. 
Suppose there is a small calibration set $\{\boldsymbol{X}^{(t)}_{j}, A_j^{(t)}, Y^{(t)}_{j}\}_{j\in \mathcal{I}_{\mathrm{label}}^{(t)}}.$
Because  only a very limited number of labeled samples are available in the target dataset (fewer than $50 $ observations in both simulations and real data), we select the hyperparameter  $\delta$  via a grid search over $[0, 1]$.  Specifically, the hyperparameter  $\delta$  is chosen to minimize  the squared prediction error:
\begin{equation}
\begin{aligned}\widehat{\delta}=\arg \min_{\delta\in [0,1]} 
   &\frac{1}{|\mathcal{I}_{\mathrm{label},1}^{(t)}|}\sum_{i\in \mathcal{I}_{\mathrm{label},1}^{(t)}} \left\{\sum_{s=1}^{|\mathcal{S}|}\widehat{\mathcal{W}}_s\left(\boldsymbol{X}_{i}^{(t)}, \widehat{\rho}^{\delta}_{n,s},\delta\right)\widehat{f}^{(s)}_{1}\left(\boldsymbol{X}_{i}^{(t)}\right)-Y^{(t)}_{i}\right\}^{2}\\
   +&\frac{1}{|\mathcal{I}_{\mathrm{label},0}^{(t)}|}\sum_{i\in \mathcal{I}_{\mathrm{label},0}^{(t)}} \left\{\sum_{s=1}^{|\mathcal{S}|}\widehat{\mathcal{W}}_s\left(\boldsymbol{X}_{i}^{(t)}, \widehat{\rho}^{\delta}_{n,s},\delta\right)\widehat{f}^{(s)}_{0}\left(\boldsymbol{X}_{i}^{(t)}\right)-Y^{(t)}_{i}\right\}^{2},
   \nonumber
   \end{aligned}
\end{equation}
where $\mathcal{I}_{\mathrm{label},1}^{(t)} = \{j \in \mathcal{I}_{\mathrm{label}}^{(t)}: A_j^{(t)} = 1\}$ and $\mathcal{I}_{\mathrm{label},0}^{(t)} = \{j \in \mathcal{I}_{\mathrm{label}}^{(t)}: A_j^{(t)} = 0\}$.
The tuned hyper-parameter $\widehat{\delta}$  makes the source-based predictions match the limited target labels as closely as possible. Finally, the PDRO-ITR  takes the form
    $$ \widehat{d}_{pdro}(\boldsymbol{X};\widehat{\delta})=\mathbb{I}\left[\sum_{s=1}^{|\mathcal{S}|}\widehat{\mathcal{W}}_s\left(\boldsymbol{X}, \widehat{\rho}_{n,s}^{\widehat{\delta}}, \widehat{\delta}\right)\widehat{C}^{(s)}\left(\boldsymbol{X}\right)>0\right].$$

\section{Theoretical Results}
We consider the worst-case  distribution   within the uncertainty $\mathcal{U}_1(\delta)$, defined as, 
$$\mathbb{T}_{Y(1), Y(0)|\boldsymbol{X}}^{0}=\arg \min_{\mathbb{T}_{Y(1), Y(0)|\boldsymbol{X}} \in \mathcal{U}_1(\delta)} \max_{d} E_{\mathbb{P}_{\boldsymbol{X}}^{(t)}}\left[C\left\{\boldsymbol{X};  \mathbb{T}_{Y(1), Y(0)|\boldsymbol{X}}\right\}d(\boldsymbol{X})\right].$$ 
Here,
$\mathbb{T}_{Y(1), Y(0)|\boldsymbol{X}}^{0}$ represents the most adversarial conditional distribution, which minimizes the policy value of the optimal ITR.
Then, define 
$$\mathcal{V}_0(d): =E_{\mathbb{P}_{\boldsymbol{X}}^{(t)}}\left[C\left\{\boldsymbol{X};  \mathbb{T}_{Y(1), Y(0)|\boldsymbol{X}}^{0}\right\}d(\boldsymbol{X})\right].$$
We aim to  establish an upper bound for  risk $\mathcal{V}_0( d^{*}_{pdro}(\boldsymbol{X}; \delta))-\mathcal{V}_0(\widehat{d}_{pdro}(\boldsymbol{X}; \delta))$.
To derive the theoretical properties of the proposed estimator, we impose the following regularity conditions.
\begin{enumerate}
  \renewcommand{\labelenumi}{(C\arabic{enumi})}
  \renewcommand{\theenumi}{C\arabic{enumi}}
 \setlength{\itemindent}{11pt} 
  \setlength{\labelsep}{5pt} 
  \setlength{\labelwidth}{0pt} 
  \setlength{\leftmargin}{0pt} %
\item   The CATE is uniformly bounded, i.e., $\big|C\left(\boldsymbol{x}; \mathbb{P}_{Y(1), Y(0)|\boldsymbol{X}}^{(s)}\right)\big| \leq c_1$, for any $\boldsymbol{x}\in \mathcal{X}$ and $s\in \mathbf{S}$.
\item  
(i) There exist  $c_2>0$ and $\alpha >0$ such that $\mathbb{P}\left[\big|f_{\boldsymbol{\rho}_0,\eta_0}^{\delta}(\boldsymbol{X})\big|  < \gamma\right]\leq c_2\gamma^{\alpha}$  for any $\gamma>0$, where $f_{\boldsymbol{\rho}_0,\eta_0}^{\delta}(\boldsymbol{X})=\sum_{s=1}^{|\mathcal{S}|}\left\{ \delta {\omega}^{(s)}_0(\boldsymbol{X})+(1-\delta)\rho_{0,s} \right\}C^{(s)}_0\left(\boldsymbol{X}\right)$; (ii) Suppose that the density function of $f_{\boldsymbol{\rho},\eta}^{\delta}(\boldsymbol{X})$ is bounded  for any $\boldsymbol{\rho}\in \mathcal{P}$ and $\eta \in \mathcal{H}$, where $\mathcal{H}=\{(\omega^{(1)},\ldots \omega^{(S)}, C^{(1)},\ldots, C^{(S)}): \omega^{(s)}\in \mathcal{F}_1,   C^{(s)} \in \mathcal{F}_0 \}$.
\item For any $s \in \mathcal{S}$ and $\nu>0$, there exist sequences $\Delta_{\omega}({n};\nu)$  and $\Delta_{C}({n};\nu)$ converging to zero as $n \rightarrow \infty$ , such that
 $E_{\mathbb{P}^{(s)}}\left\{\widehat{C}^{(s)}\left(\boldsymbol{X}\right)-C\left(\boldsymbol{X};\mathbb{P}_{Y(1), Y(0)|\boldsymbol{X}}^{(s)}\right)\right\}^{2}
\leq \Delta_{C}({n};\nu)$ and  $E_{\mathbb{P}^{(s)}}
\left\{\widehat{\omega}^{(s)}(\boldsymbol{X})-\omega^{(s)}_{0}(\boldsymbol{X})\right\}^{2}\leq \Delta_{\omega}({n};\nu)$ with probability at least $1-\nu/2$.

\item The propensity score function $P(A^{(s)} = a |\boldsymbol{X}^{(s)}= \boldsymbol{x}) \geq c_4 > 0$, for $a \in \mathcal{A}$, $\boldsymbol{x} \in \mathcal{X}$ and $s \in \mathcal{S}$.

\item  
Let $\boldsymbol{C}_{0}(\boldsymbol{x})=\left(C\left(\boldsymbol{x};\mathbb{P}_{Y(1), Y(0)|\boldsymbol{X}}^{(1)}\right),  C\left(\boldsymbol{x};\mathbb{P}_{Y(1), Y(0)|\boldsymbol{X}}^{(2)}\right), \ldots, C\left(\boldsymbol{x};\mathbb{P}_{Y(1), Y(0)|\boldsymbol{X}}^{|\mathcal{S}|}\right) \right)^{\!\top}$.
Assume  that,  for any $\boldsymbol{\rho}\in \mathcal{P} $ and $\delta \in [0, 1]$,  the matrix
\[
E_{\mathbb{P}_{X}^{(t)}}\left[\boldsymbol{C}_0(\boldsymbol{X})\boldsymbol{C}_0^{\!\top}(\boldsymbol{X})\,\mathbb{I}\left\{\bigg|\sum_{s=1}^{|\mathcal{S}|}\mathcal{W}(\boldsymbol{X}, \rho_{s},\delta)C\left(\boldsymbol{X};  \mathbb{P}_{Y(1),Y(0)|\boldsymbol{X}}^{(s)}\right)\bigg|<h\right\}\right], 
\] has minimal eigenvalue $\lambda_h \geq c_3 h^{\beta} >0$ and $0<\beta<3$.

\item Assume that  $f \in  \mathcal{F}_0$ satisfies $|f(\boldsymbol{x})| \leq \mathcal{B}_0/2$ for  any $\boldsymbol{x}\in \mathcal{X}$, where  $ \mathcal{F}_0$ is the  class of  FNNs. 
\end{enumerate}

Condition (C1) holds if the treatment effect $Y(1)-Y(0)$ is uniformly bounded. Condition (C2) (i) is a margin condition commonly used in \cite{qian2011performance} and \cite{shi2020breaking}; Condition (C2) (ii) means that $\alpha \geq 1$. Condition (C3) 
specifies the error rates of the general machine learning estimators for the nuisance parameters.  
Condition (C4) ensures sufficient overlap between treatment groups. Condition (C5) ensures  identifiability of the weights $\boldsymbol{\rho}^{\delta}_{0}$. Actually, the matrix in  Condition (C5)  is  positive semi-definite. Condition (C6) imposes boundedness on the FNNs class,  which is standard condition in DNNs literature  \citep{jiao2023deep, yan2025deep}.

The following lemma characterizes the relationship between the risk and the estimation error associated with the nuisance parameter.

\begin{lemma}
Suppose that Conditions {(C1)-(C4)} hold. For each $\delta \in [0,1]$, $h > 0$ and $\nu>0$,
we have
\begin{equation}
\label{lemma_1}
\begin{aligned}
&\mathcal{V}_0( d^{*}_{pdro}(\boldsymbol{X}; \delta))-\mathcal{V}_0(\widehat{d}_{pdro}(\boldsymbol{X}))\\
\lesssim &
\left[\delta^{2}\Delta_{\omega}({n};\nu)+(1-\delta)^{2}\sum_{s=1}^{|\mathcal{S}|} (
\rho^{\delta}_{0,s}-\widehat{\rho}^{\delta}_{n,s})^{2}
+\Delta_{C}({n};\nu)\right]^{\frac{1+\alpha}{2+\alpha}}, 
\end{aligned}
\end{equation}
$\text{ with probability at least } 1-\nu,$
where $a \lesssim  b$ denotes $a \leq cb$ for a constant $c>0$. 
\end{lemma}
By Lemma 1, the risk is controlled by $\Delta_{\omega}({n};\nu)$ 
and $\Delta_{C}({n};\nu)$ when $\delta=1$,
whereas the risk is dominated by  $\sum_{s=1}^{|\mathcal{S}|} (
\rho^{\delta}_{0,s}-\widehat{\rho}^{\delta}_{n,s})^{2}$ and $\Delta_{C}({n};\nu)$  when $\delta=0$.
Building on this decomposition, the following theorem establishes an upper bound on the  risk under additional regularity conditions.
\begin{theorem}
Suppose that Conditions {(C1)-(C6)} hold. For each $\delta \in [0,1)$, $h>0$ and $\nu>0$,
we have
\begin{equation}
\label{thm_3_value_decomposition}
\begin{aligned}
&\mathcal{V}_0\big( d^{*}_{pdro}(\boldsymbol{X}; \delta)\big) - \mathcal{V}_0\big(\widehat{d}_{pdro}(\boldsymbol{X}; \delta)\big) \\
&\lesssim 
\left[\Delta_{\omega}({n};\nu)+\Delta_{C}({n};\nu)+h^{2-2\beta}n^{-1}+h^{3-\beta}
\right]^{\frac{1+\alpha}{2+\alpha}}, \text{with probability at least } 1-2\nu.
\end{aligned}
\end{equation}
\end{theorem}{
Under the standard regularity conditions for multinomial logistic regression \citep{fox2015applied, wu2025class}, the  mean squared error between $\widehat{\omega}^{(s)}(\boldsymbol{X})$ and $\omega^{(s)}_{0}(\boldsymbol{X})$ has  convergence  rate $\Delta_{\omega}({n};\nu)=O_p(n^{-1})$. For the CATE estimators obtained via feedforward neural networks, we have the convergence rate  $\Delta_{C}(n; \nu)=O_p\left(n^{-2\tau/(p+2\tau)}\right)$ under suitable regularity conditions \citep{jiao2023deep}, where $\tau$ denotes the smoothness parameter of Hölder  class.
In particular, when $\alpha=0$, $\beta=1$, and $h=O(n^{-1/2})$, it follows that  $\mathcal{V}_0\big( d^{*}_{pdro}(\boldsymbol{X}; \delta)\big) - \mathcal{V}_0\big(\widehat{d}_{pdro}(\boldsymbol{X}; \delta)\big)=O_p\left( n^{-\tau/(p+2\tau)}+n^{-1/2}\right)$.}

\section{Simulation Studies}

\subsection{Simulation Designs}
We conducted  simulation studies in four scenarios to evaluate the performance of the proposed PDRO-ITR. 
In each scenario, we considered three source groups. 

For each group $S=s$, the treatment variable $A^{(s)}$ was randomly assigned from $\{0,1\}$ with equal probability. 
Each component of $\boldsymbol{X}^{(s)}$ was independently sampled from $\mathcal{N}(0,1)$ and truncated to the interval $[-10, 10]$.  
The group assignment probability was given by
$\omega_{0}^{(s)}( \boldsymbol{x})=P(S=s|\boldsymbol{X}=\boldsymbol{x})= {\exp(\boldsymbol{\beta}_s^{T}\boldsymbol{x})}/\{\sum_{j=1}^3 \exp(\boldsymbol{\beta}_j^{T}\boldsymbol{x})\}$ with  $\boldsymbol{\beta}_1^{T}=(-3, 2 ,1, 0, 0),\boldsymbol{\beta}_2^{T}=(1, -1, 3, 0, -1) $ and $\boldsymbol{\beta}_3^{T}=(1, 0 ,0, -1, 2)$.
The source outcomes were generated from the model
\begin{equation}
\begin{aligned}
Y^{(s)} &= f^{(s)}(\boldsymbol{X}^{(s)}) (2A^{(s)}-1) + \varepsilon^{(s)},
\end{aligned}
\label{eq:experiment-definitions}
\end{equation}
where the noise terms $\epsilon^{(s)}  \sim \mathcal{N}(0,1)$ was independent of $\boldsymbol{X}^{(s)}$ for $s=1, 2, 3$.

The target outcome was generated by
\begin{equation}
\begin{aligned}
Y^{(t)} &= \left[\delta \sum_{s=1}^3 \omega_{0}^{(s)}( \boldsymbol{X}^{(t)}) f^{(s)}(\boldsymbol{X}^{(t)}) + (1-\delta)  \sum_{s=1}^3 \rho_s f^{(s)}(\boldsymbol{X}^{(t)})\right] (2A^{(t)}-1)+ \varepsilon^{(t)},
\end{aligned}
\label{eq:experiment-definitions}
\end{equation}
where $\varepsilon^{(t)} \sim \mathcal{N}(0,1)$ was independent of $\boldsymbol{X}^{(t)}$, and $A^{(t)}$  was randomly assigned from $\{0,1\}$ with equal probability. The components of covariates $\boldsymbol{X}^{(t)}$ were generated independently from a standard normal distribution and truncated to the interval $[-10, 10]$.

The dimensions of $\boldsymbol{X}^{(s)}$ and $\boldsymbol{X}^{(t)}$ were $5,5,30,$ and $30$ for Scenarios 1-4, respectively. For the four scenarios, the functions $\{f^{(s)}\}_{s=1}^{3}$ were specified as follows:

\noindent\textbf{Scenario 1:}
\begin{equation}
\begin{aligned}
f^{(1)}(\boldsymbol{x}) &= 3x_1 + x_3 + x_4 - x_5, \\
f^{(2)}(\boldsymbol{x}) &= x_1- x_2 - 2x_3 +x_4 + x_5, \\
f^{(3)}(\boldsymbol{x}) &= x_1 + 2x_2 + x_3 - x_4 + x_5;
\nonumber
\end{aligned}
\end{equation}
\noindent\textbf{Scenario 2:}
\begin{equation}
\begin{aligned}
f^{(1)}(\boldsymbol{x}) &= -\sin(x_1) + \exp\left(\frac{x_2}{10}\right) - (x_3 - x_4)^2 + (x_5)^3, \\
f^{(2)}(\boldsymbol{x}) &= \sin(x_1) - x_2 x_3 - (x_3)^2 + (x_4)^2 - \max(0, x_5), \\
f^{(3)}(\boldsymbol{x}) &= -2 x_1 - (x_2)^2 + (x_3)^2 - x_4 + |x_5|;
\nonumber
\end{aligned}
\label{eq:case2-f}
\end{equation}
\noindent\textbf{Scenario 3:}
\begin{equation}
\begin{aligned}
f^{(1)}(\boldsymbol{x}) &= x_1 + x_2 +x_3 + x_4 - 3x_5, \\
f^{(2)}(\boldsymbol{x}) &= x_1- 2x_2 + 2x_3 +x_4 + 3x_5, \\
f^{(3)}(\boldsymbol{x}) &= x_1 + x_2 + x_3 - x_4;
\nonumber
\end{aligned}
\end{equation}
\noindent\textbf{Scenario 4:}
\begin{equation}
\begin{aligned}
f^{(1)}(\boldsymbol{x}) &=
\sin(x_{1}) + \exp(x_{2}+x_{3}) + (x_{4}-3x_{5})^2 + 3x_{6}, \\
f^{(2)}(\boldsymbol{x}) &=\max\{0, x_{1} x_{2}\} + x_{3} - x_{4} + x_{5}^2, \\
f^{(3)}(\boldsymbol{x}) &=-5x_{1} - x_{2}^3 - (x_{3}-x_{4})^2 + |x_{5}|.
\nonumber
\end{aligned}
\end{equation}

The source dataset consists of $n$ samples, with $N_s$ samples from the $s$-th source. We generated $N_t=25$ samples as a calibration set to tune hyper-parameter $\delta$. We also generated $\{\boldsymbol{X}_{i}^{(t)}\}_{i=1}^{1000}$ as a test set to compute the empirical policy value.
For any estimated ITR $\widehat{d}$, its empirical policy value was defined as
$$
\frac{1}{1000}\sum_{i=1}^{1000} C\left\{\boldsymbol{X}_{i}^{(t)}; \mathbb{P}_{Y(1), Y(0)|\boldsymbol{X}}^{(t)}\right\} 
   \widehat{d}(\boldsymbol{X}_{i}^{(t)}).
$$
To evaluate distributional robustness, we sampled $100$ different $\boldsymbol{\rho}$ vectors from a Dirichlet distribution and report the worst-case policy value  over the $100$ trials.  We compared our method with the following methods.

\textbf{Naive:} The CATE was estimated separately for each source (see Section 3.1, Step 1), and the ITR was constructed by  the sample-weighted average CATE, 
\[
\widehat{d}(\boldsymbol{x}) = \mathbb{I}\Biggl[\sum_{s \in\mathcal{S}} \frac{N_s}{\sum_{j \in\mathcal{S}} N_j} \widehat{C}^s(\boldsymbol{x}) > 0 \Biggr].
\]

\textbf{MR-CATE:}  \citet{zhang2024minimax} proposed a minimax regret estimator for CATE using multiple source data. Then, the corresponding ITR is  derived by the CATE estimator.

\textbf{MPL:}   \citet{shi2018maximin}  introduced a maximin projection learning method for optimal ITR.

\textbf{DRO:}  The ITR estimated by solving the distributionally robust optimization problem in Equation \eqref{e_dro}.

\subsection{Simulation Results}
Figures 3 and 4 demonstrate that the proposed PDRO-ITR method consistently attains the highest policy value across different scenarios and $\delta$, outperforming existing approaches. In particular, as $\delta$ approaches $1$, the target model incorporates more distributional shift induced by $\omega_{0}^{(s)}$, under which the PDRO-ITR exhibits notably superior performance.
When $\delta$ is close to $0$, the distributional shift is primarily governed by $\{\rho_s\}_{s\in \mathcal{S}}$, and the PDRO-ITR method exhibits competitive performance relative to DRO-ITR.   These results indicate that the proposed method performs robustly under varying levels of distributional shift.

\begin{figure}[h]
    \centering
    \begin{minipage}[b]{0.49\linewidth}
        \centering
        \includegraphics[width=\linewidth]{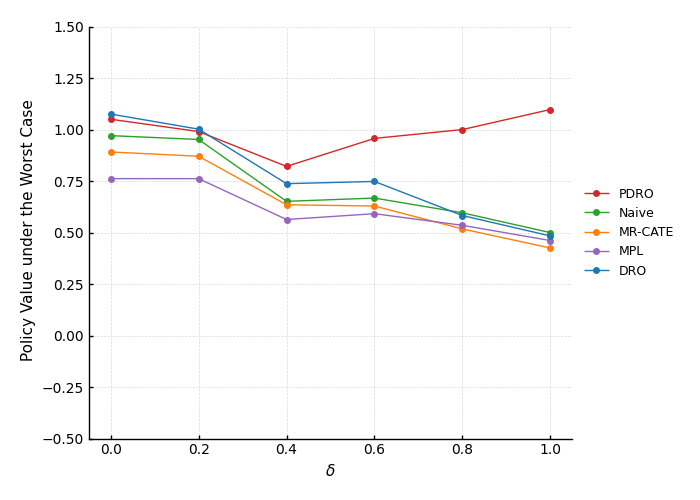}
 \parbox{1\linewidth}{\centering (a) Scenario 1}
    \end{minipage}
    \hfill
    \begin{minipage}[b]{0.49\linewidth}
        \centering
        \includegraphics[width=\linewidth]{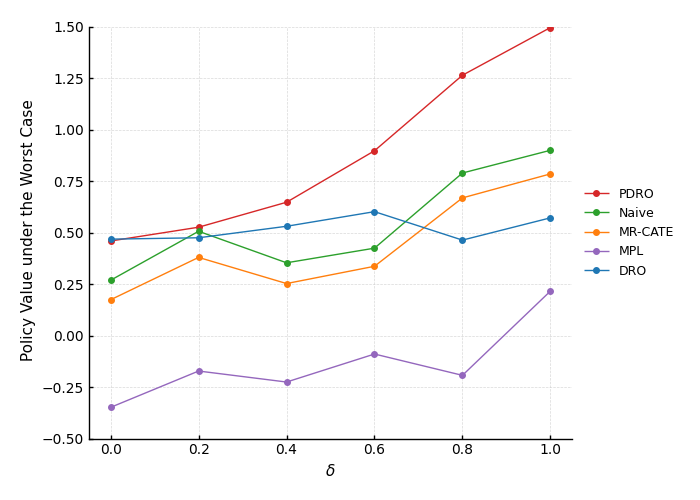}
 \parbox{1\linewidth}{\centering (b) Scenario 2}
    \end{minipage}
    \caption{Policy value plots under the worst-case for Scenarios 1 and 2, with varying $\delta$,  $n=2000$, and $200$ repetitions.}
\end{figure}

\begin{figure}[h]
    \centering
    \begin{minipage}[b]{0.49\linewidth}
        \centering
        \includegraphics[width=\linewidth]{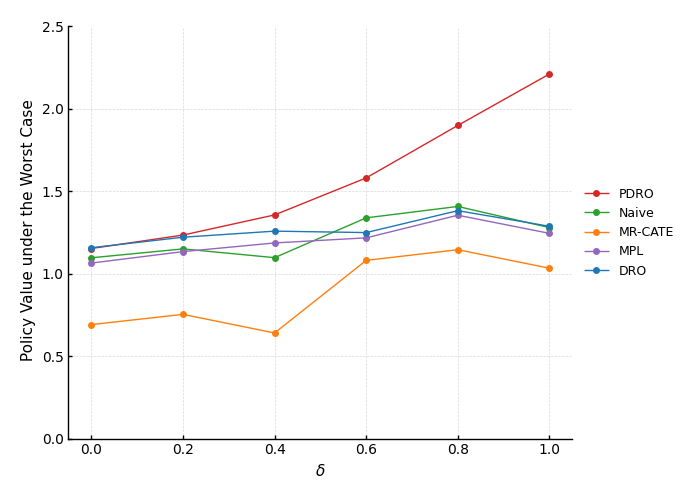}
     \parbox{1\linewidth}{\centering (a) Scenario 3}
    \end{minipage}
    \hfill
    \begin{minipage}[b]{0.49\linewidth}
        \centering
        \includegraphics[width=\linewidth]{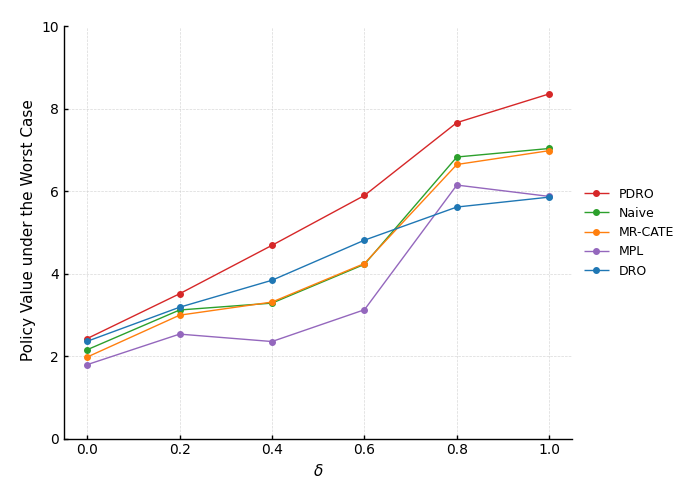}
  \parbox{1\linewidth}{\centering (b) Scenario 4}
    \end{minipage}
    \caption{Policy value plots under the worst-case for Scenarios 3 and 4, with varying $\delta$,  $n=2000$, and $200$ repetitions.}
    \label{fig:policy-value}
\end{figure}

In addition, the proposed DRO-ITR method  outperforms MR-CATE method when $\delta$ is close to $0$. This improvement arises because DRO-ITR directly maximizes the policy value under the worst-case to estimate the parameter $\rho^{*}$, whereas MR-CATE minimizes the worst-case regret. Nonetheless, the two objective functions are not equivalent.

Furthermore, PDRO-ITR remains effective in both linear (Scenarios 1 and 3) and nonlinear (Scenarios 2 and 4) settings. In contrast, the MPL method shows inferior performance in nonlinear scenarios, as it is restricted to a linear decision function class.

\begin{table*}[h]
\caption{Sample means and sample standard deviations (in parentheses) of  policy value (PV) for the proposed PDRO-ITR, DRO-ITR, Naive, MR-CATE and MPL methods, based on 200 replications under Scenarios 1 and 2 with $\delta=0.75$.}
\centering
\begin{tabular}{cccccc}
\toprule
 & Methods &$n=500$  & $n=1000$  & $n=2000$ \\
\midrule
\multirow{4}{*}{Scenario 1} 
& PDRO  &   \textbf{0.984} (0.028)  &  \textbf{1.008} (0.018) &   \textbf{1.017} (0.012) \\
& Naive  & 0.733(0.051)     &0.719  (0.034) & 0.695 (0.028) \\
& MR-CATE  &0.613 (0.079)  &  0.596 (0.061)  & 0.575 (0.055)\\
& MPL   &0.590 (0.078)  &  0.601 (0.058)  &0.608 (0.047)\\
& DRO  &  0.655 (0.117)  &0.674  (0.064)   &0.629 (0.069)\\
\midrule
\multirow{4}{*}{Scenario 2} 
& PDRO   &  \textbf{0.904}  (0.115) &    \textbf{1.038}  (0.049) &   \textbf{1.095}  (0.020) \\
& Naive       & 0.524  (0.113) & 0.600  (0.052) & 0.646  (0.033) \\
& MR-CATE    &0.382  (0.134) & 0.475  (0.067) &0.523  (0.033)\\
& MPL    &  0.094  (0.110) &  0.065  (0.093) &0.069  (0.073)\\
&DRO    &    0.429  (0.152)           &      0.526  (0.101)         &0.562  (0.084)\\
\bottomrule
\end{tabular}
\end{table*}

Tables 1 and 2 indicate that the proposed PDRO-ITR method consistently attains the highest policy values across different sample sizes. Moreover, PDRO-ITR achieves the smallest standard deviations among competing methods in Scenarios 1, 2, and 3. In Scenario 4, its standard deviation is the second lowest, except for the biased MPL estimator, whose policy value is substantially lower than that of PDRO-ITR. These results demonstrate the stability and robustness of PDRO-ITR to sampling variability.

\begin{table*}[h]
\caption{Sample means and sample standard deviations (in parentheses) of  policy value (PV) for the proposed PDRO-ITR, DRO-ITR, Naive, MR-CATE and MPL methods, based on 200 replications under Scenarios 3 and 4 with $\delta=0.75$.}
\centering
\begin{tabular}{cccccc}
\toprule
 & Methods &$n=500$  & $n=1000$  & $n=2000$ \\
\midrule
\multirow{4}{*}{Scenario 3} 
& PDRO  &   \textbf{1.607} (0.149) & \textbf{1.826} (0.024)&\textbf{1.879}  (0.015) \\
& Naive  & 1.176  (0.133)   &1.331  (0.047) & 1.357  (0.032) \\
& MR-CATE  &0.949  (0.092) &  1.015  (0.045)  &1.027  (0.035)\\
& MPL   &1.256  (0.069)  &  1.307  (0.045)  &1.316  (0.035)\\
& DRO  &  1.024  (0.287)  &1.293  (0.094)   &1.354  (0.041)\\
\midrule
\multirow{4}{*}{Scenario 4} 
& PDRO   & \textbf{6.243}  (0.195)  &      \textbf{6.480} (0.147)  &    \textbf{6.714}    (0.091)   \\
& Naive       & 4.793  (0.165) & 5.066  (0.208) & 5.504  (0.201) \\
& MR-CATE    & 4.849  (0.220) & 5.097  (0.209)  &5.446  (0.185)\\
& MPL    & 4.641  (0.019) &  4.635  (0.001) &4.635  (0.000)\\
&DRO    &  1.698  (1.228)         & 4.170  (0.667)    &4.919  (0.325)\\
\bottomrule
\end{tabular}
\end{table*}

\section{Applications}
We evaluate the generalizability of the proposed method on two real-world datasets: the AIDS Clinical Trials Group Study 175 (ACTG)\citep{hammer1996trial} and the Oregon Health Insurance Experiment (OHIE) \citep{finkelstein2012oregon}. The OHIE data are available at {https://www.nber.org/research/data/oregon-health-insurance-experiment-data}, and the ACTG data are available at   
{https://archive.ics.uci.edu/dataset/890/aids+clinical+trials}.

\subsection{The AIDS Clinical Trials Group Study}
ACTG was a randomized, double-blind clinical study designed to evaluate four treatments  for adults infected with human immunodeficiency virus type 1 (HIV-1) whose CD4 cell counts were between 200 and 500 per cubic millimeter \citep{hammer1996trial}. A total of $2467$ HIV-1–infected patients were randomly assigned to one of four daily treatment arms: 600 mg of zidovudine (ZDV), 600 mg of ZDV plus 400 mg of didanosine (ZDV + ddI), 600 mg of ZDV plus 2.25 mg of zalcitabine (ZDV + ZAL), or 400 mg of  didanosine (ddI). In our analysis, we considered the outcome $Y$ as the change in CD4 cell count from baseline to the early follow-up stage (20 ± 5 weeks).  We focus on a binary treatment comparison between ZDV + ZAL ($A=1$) and ddI ($A=-1$), including $1085$ patients.

It is worth noting that AIDS clinical research faces generalizability concerns.  In particular, restrictive eligibility criteria frequently exclude large proportions of HIV-infected women \citep{gandhi2005eligibility}. In our analysis of the ACTG  dataset, only $72$ participants  out of $1085$ were  white females. We therefore treat white females as the target population, aiming to  improve the performance of the proposed ITR for this subgroup. We used $50$ target samples to tune hyperparameters. Participants from other demographic groups were considered as source data. The overall population structure was summarized in Table \ref{table3}.
\begin{table*}[h]
\caption{Population structure for  ACTG data.}
\centering
\begin{tabular}{ccccc}
\toprule
 &White male&  Minority male &  Minority female &  White female \\
\midrule
Sample size &690 & 215   & 108 & 72\\
\bottomrule
\label{table3}
\end{tabular}
\end{table*}
To evaluate the performance of the proposed PDRO-ITR, we used the doubly robust empirical policy value estimator \citep{zhang2012robust,shi2020breaking}. Specifically, for any ITR $d$, its empirical policy value was given by
{\small $$\frac{1}{N_t}\sum_{i=1}^{N_t} \left[\frac{\mathbb{I}\{A_i=d(\boldsymbol{X}_i)\}}{\pi(A_i,\boldsymbol{X}_i)}\{Y_i-\sum_{a=0,1}\widehat{f}_a(\boldsymbol{X}_i)\mathbb{I}(A_i=a)\}+d(\boldsymbol{X}_i)\widehat{f}_1(\boldsymbol{X}_i)+\{1-d(\boldsymbol{X}_i)\}\widehat{f}_0(\boldsymbol{X}_i)\right],$$}
where $\boldsymbol{X}_i$ $(i=1,2,\ldots,N_t)$ were the target samples and the propensity score was $\pi(a,\boldsymbol{x}) = P(A=a\mid \boldsymbol{X}=\boldsymbol{x}) \equiv 0.5$ in the ACTG randomized trial. The  functions $\widehat{f}_a(\boldsymbol{x})=E(Y^{(t)}|\boldsymbol{X}^{(t)}=\boldsymbol{x}, A^{(t)}=a)$, $a=0,1$, were estimated using the procedure described in Step 1 of Section 4.  The covariates included  age, weight, hemophilia, homosexual activity, history of IV drug use, Karnofsky score,  antiretroviral history, antiretroviral history stratification, symptomatic indicator, baseline CD4 count  and baseline CD8 count.

As shown in Table \ref{table4}, the proposed PDRO-ITR method attains a higher policy value than all comparison methods.
\begin{table*}[h]
\caption{Policy values for the target group across five methods on  ACTG data.}
\centering
\begin{tabular}{lcccccc}
\toprule
Method       & PDRO   & Naive   & MR-CATE     & MPL    & DRO  \\
\midrule
Policy value & 31.519    & 22.569  &   25.338  & 28.459   & 29.200 \\
\bottomrule
\label{table4}
\end{tabular}
\end{table*}

\subsection{Oregon Health Insurance Experiment}
In 2008, the state of Oregon implemented a lottery to expand Medicaid coverage among low-income, uninsured adults. Individuals selected in the lottery were allowed to apply for Medicaid, which provided comprehensive healthcare coverage, including physician services, prescription medications and mental health care. Twelve months after enrollment, a mail survey was administered to assess multiple outcomes.

In our analysis, the treatment $A=1$ denoted receipt of free healthcare through the Medicaid program. The outcome was the score on the physical component of the 8-item short form health survey, where higher scores indicate better self-reported physical health.
The covariate vector included 16 individual-level characteristics. Medical history variables comprised the number of emergency room visits prior to the experiment, self-reported overall health,  indicators for chronic conditions and others. Demographic covariates included age, income, and education level.

\begin{table*}[h]
\caption{Population structure for  OHIE data.}
\centering
\begin{tabular}{ccccc}
\toprule
  White & Hispanic & Black  & Asian & Other  \\
\midrule
  3565 & 620 & 392  & 362 & 730\\
\bottomrule
\end{tabular}
\end{table*}
After excluding observations with missing data, the analytic sample consisted of 5309 individuals. Participants identifying as White, Hispanic, Black, or Asian were treated as the source population, and those identifying with other racial or ethnic categories as the target population, where the total population structure is shown in Table 5. The policy value was estimated using the double robust procedure described in Section 7.1. The propensity score,
$\pi(a,\boldsymbol{x}) = P(A=a \mid \boldsymbol{X}=\boldsymbol{x})$,
was estimated via logistic regression \citep{yu2011dual}. Hyper-parameter $\delta$ were tuned using 50 target samples.
Table 6 reports that the proposed PDRO-ITR method achieves the highest estimated policy value than existing approaches.
\begin{table*}[h]
\caption{Policy values for the target group across five methods on  OHIE data.}
\centering
\begin{tabular}{lcccccc}
\toprule
Method       & PDRO   & Naive   & MR-CATE     & MPL    & DRO  \\
\midrule
Policy value & 49.750    & 49.432 &49.338     & 49.181  &48.833\\
\bottomrule
\end{tabular}
\end{table*}

\section{Discussion}
We proposed  a novel approach for estimating individualized treatment rules  that are robust to conditional distribution shift. The proposed  PDRO-ITR leverages multi-source information to construct a covariate-dependent uncertainty set, balancing robustness and efficiency through the mixing parameter~$\delta$. The framework admits a closed-form characterization, avoiding the need to solve the original max–min optimization directly, and achieves competitive performance in both simulation and real-data analyses.

Several promising directions for future research are worth further investigation.
First,  a natural extension is to  simultaneously consider both  covariates shift and posterior shift. 
Define the density ratio 
$r(\boldsymbol{x}) = d\mathbb{P}^{(t)}_{\boldsymbol{X}}(\boldsymbol{x}) / d\mathbb{P}^{(mix)}_{\boldsymbol{X}}(\boldsymbol{x})$, 
where $d\mathbb{P}^{(t)}_{\boldsymbol{X}}(\boldsymbol{x})$ denotes the density function of the target covariates $\boldsymbol{X}^{(t)}$  and 
$d\mathbb{P}^{(mix)}_{\boldsymbol{X}}(\boldsymbol{x})$ represents the density function of a mixture distribution over the multiple sources covariates. 
Then, the policy value function can be written as
\begin{equation*}
E_{\mathbb{P}_{\boldsymbol{X}}^{(t)}}\!\left[C\!\left\{\boldsymbol{X}; \mathbb{P}_{Y(1), Y(0)|\boldsymbol{X}}^{(t)}\right\}d(\boldsymbol{X})\right]
= 
E_{\mathbb{P}_{\boldsymbol{X}}^{(mix)}}\!\left[r(\boldsymbol{X})\, C\!\left\{\boldsymbol{X}; \mathbb{P}_{Y(1), Y(0)|\boldsymbol{X}}^{(t)}\right\}d(\boldsymbol{X})\right].
\end{equation*}
Accordingly, a PDRO-ITR under covariate shift can be obtained 
\begin{equation}
\underset{d}{\max}\;
\underset{\mathbb{T}_{Y(1), Y(0)|\boldsymbol{X}} \in \mathcal{U}_{1}(\delta)}{\min}\;
E_{\mathbb{P}_{\boldsymbol{X}}^{(mix)}}\!\left[r(\boldsymbol{X})\, C\!\left\{\boldsymbol{X}; \mathbb{T}_{Y(1), Y(0)|\boldsymbol{X}}\right\}d(\boldsymbol{X})\right],
\label{eq:pdro_covshift}
\end{equation}
where a variety of methods can be employed to estimate $r(\boldsymbol{x})$, such as recent deep learning–based density ratio estimators \citep{xu2025estimating}. 
Another promising direction is to extend the PDRO-ITR to {dynamic treatment regimes}, enabling robust decision-making in sequential settings.
{Finally, developing formal statistical tests for distributional shift remains an important direction for future research. If such a shift is detected, it is important to quantify its magnitude and to identify which source distributions can be leveraged to improve estimation in the target population}

\section*{Supplementary Materials}
The supplementary materials include the proofs of Lemma 1 and Theorems 1-3.

\bibliographystyle{chicago}
\bibliography{reference}

\end{document}